\documentclass{article}

     \usepackage[final]{neurips_2019}

\usepackage[utf8]{inputenc} 
\usepackage[T1]{fontenc}    
\usepackage{hyperref}       
\usepackage{url}            
\usepackage{booktabs}       
\usepackage{amsfonts}       
\usepackage{nicefrac}       
\usepackage{microtype}      

\usepackage{my_pkg}

\title{Torchreid: A Library for Deep Learning Person Re-Identification in Pytorch}

\author{%
  Kaiyang Zhou \quad Tao Xiang \\
  University of Surrey, UK \\
  \\
  \url{https://github.com/KaiyangZhou/deep-person-reid} \\
}

\begin{document}

\maketitle

\begin{abstract}
Person re-identification (re-ID), which aims to re-identify people across different camera views, has been significantly advanced by deep learning in recent years, particularly with convolutional neural networks (CNNs). In this paper, we present \texttt{Torchreid}, a software library built on PyTorch that allows fast development and end-to-end training and evaluation of deep re-ID models. As a general-purpose framework for person re-ID research, Torchreid provides (1) unified data loaders that support 15 commonly used re-ID benchmark datasets covering both image and video domains, (2) streamlined pipelines for quick development and benchmarking of deep re-ID models, and (3) implementations of the latest re-ID CNN architectures along with their pre-trained models to facilitate reproducibility as well as future research. With a high-level modularity in its design, Torchreid offers a great flexibility to allow easy extension to new datasets, CNN models and loss functions.
\end{abstract}

\section{Introduction}

\begin{wrapfigure}{r}{0.5\textwidth}
\centering
\includegraphics[width=0.49\textwidth]{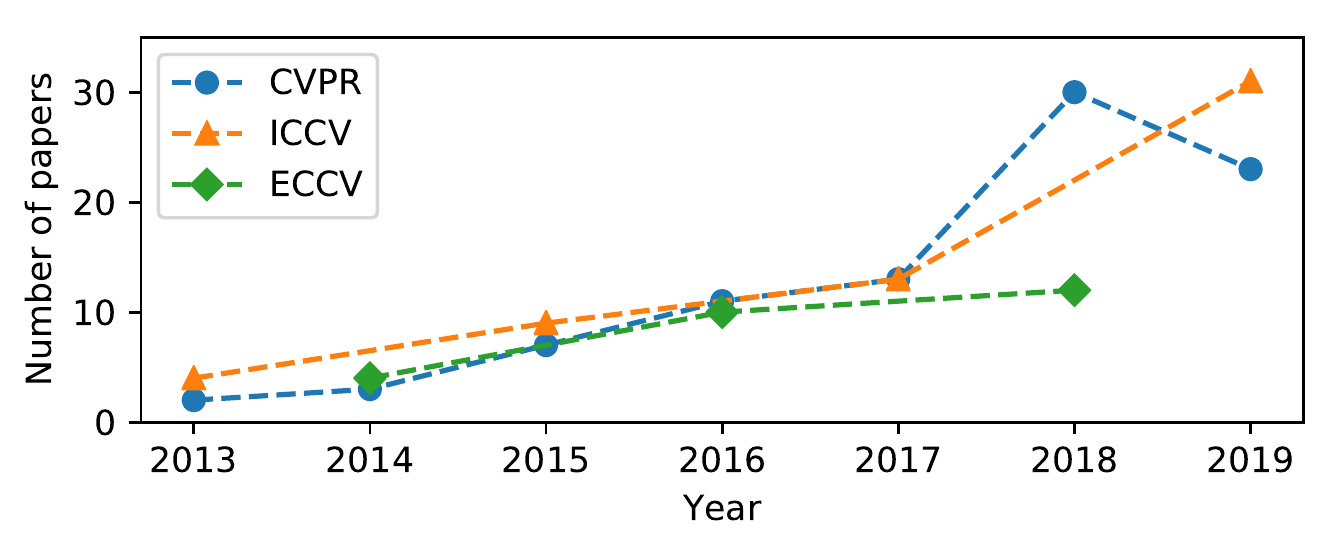}
\caption{Number of papers on person re-ID that are published in top-tier computer vision conferences over 2013 - 2019.}
\label{fig:paper_num_curve}
\end{wrapfigure}

Driven by the growing demands for intelligent surveillance and forensic applications, person re-identification (re-ID) has become a topical research area in computer vision. This is evidenced by the increasing amount of research papers published in top-tier computer vision venues in recent years (see Figure~\ref{fig:paper_num_curve}). In particular, by digging into the title and abstract of the papers, we can observe a general trend that person re-ID research has moved from feature engineering~\citep{liao2015person,matsukawa2016hierarchical} and metric learning~\citep{liao2015person,zhang2016learning}, a two-stage pipeline, to end-to-end feature representation learning with deep neural networks~\citep{li2014deepreid,ahmed2015improved,li2018harmonious,chang2018multi,zhou2019osnet,chen2019mixed,hou2019interaction}, particularly convolutional neural networks (CNNs). This can be attributed to the rapid advancement of deep learning technology, e.g., network architectures~\citep{he2016deep,xie2017aggregated,huang2017densely,sandler2018mobilenetv2,zhang2018shufflenet}, optimisation/training techniques~\citep{kingma2014adam,reddi2018on,ioffe2015batch,cosineLR,Dropout,liu2019radam}, as well as to the open-source deep learning frameworks, such as Caffe~\citep{jia2014caffe}, PyTorch~\citep{paszke2017automatic}, TensorFlow~\citep{abadi2016tensorflow} and MXNet~\citep{chen2015mxnet}, which enable researchers to quickly implement and test ideas and develop their own deep-learning projects.

Although the source code of some deep re-ID models has been released to the public, they typically differ in programming languages, backend frameworks, data-loading components, evaluation procedures, etc., which make the comparison between different approaches more difficult. This becomes more obvious when one wants to adapt the code of a published method to a new dataset or task for comparison, which may require a considerable amount of time spent on understanding and modifying the source code. This motivates us to design a generic framework that provides a standardised data-loading interface, basic training pipelines compatible with different re-ID models, and more importantly, is easy to extend.

Over the past decade, more than 30 person re-ID benchmark datasets have been introduced to the re-ID community\footnote{See \url{https://github.com/NEU-Gou/awesome-reid-dataset} for a nice summary.}, evolving from small datasets like VIPeR~\citep{gray2007evaluating} and GRID~\citep{loy2009multi} (with around thousands of images) to big datasets such as MSMT17~\citep{wei2018person} (with over 100k images). However, even the largest re-ID dataset to date is still considered as being of moderate size when compared with contemporary large-scale datasets such as ImageNet~\citep{deng2009imagenet}. This is mainly due to the difficulty and expensive cost in collecting person images with pair-wise annotations across disjoint camera views. Therefore, it is common to combine different re-ID datasets for CNN model training~\citep{xiao2016learning} or pre-training (typically followed by fine-tuning on small datasets~\citep{li2017person,liu2017hydraplus,zhao2017spindle,zhou2019osnet}). From an engineering perspective, this requires the data loaders to accept an arbitrary number of training datasets and automatically adjust the identity and camera view labels to avoid conflict. Moreover, to allow evaluation of cross-dataset performance, the data loaders also need to be able to process test images from different datasets.

In this paper, we present \texttt{Torchreid}, which is a software library built on PyTorch to provide not only a unified interface for both image and video re-ID datasets, but also streamlined pipelines that allow fast development and end-to-end training and evaluation of deep re-ID models. Specifically, Torchreid supports 15 commonly used re-ID datasets, including 11 image datasets and 4 video datasets. Basically, users are allowed to select any number of (available) datasets of the same type (image or video) for model training and evaluation.

For CNN model learning, Torchreid currently implements two training pipelines, which are classification with \texttt{softmax}\footnote{Cross-entropy loss.} loss and metric learning with \texttt{triplet}\footnote{Hard example mining triplet loss~\citep{hermans2017defense}.} loss, the two widely used (and most effective) objective functions in the literature. Users are allowed to choose either one of them or a weighted combination. Furthermore, Torchreid contains implementations of the latest state-of-the-art re-ID CNNs, together with their pre-trained models publicly available in Torchreid's model zoo\footnote{Torchreid's model zoo: \url{https://kaiyangzhou.github.io/deep-person-reid/MODEL_ZOO}.}. Therefore, with Torchreid one can quickly get hands-on experience to train a strong baseline model or build on the current state of the arts for further research. More importantly, following the principle of code reusability and structural modularity, the training pipelines are carefully designed and structured such that a new pipeline can be efficiently constructed without much code re-writing.

Besides the efficient data loading, training and evaluation procedures, Torchreid also provides a visualisation toolkit to aid the understanding of re-ID model learning, e.g., visualisation of ranking results and activation maps, as well as a full documentation\footnote{Torchreid's documentation: \url{https://kaiyangzhou.github.io/deep-person-reid/.}} including tutorials, how-to instructions, package references, etc. to help users quickly get familiar with the library.

The rest of the paper is organised as follows. Section~\ref{sec:overview} gives an overview of the Torchreid library. Section~\ref{sec:main_modules} introduces Torchreid's main modules. Section~\ref{sec:discussion} concludes this paper with a discussion.

\section{Overview of Torchreid} \label{sec:overview}
Torchreid is a library specifically developed for deep learning and person re-ID research, with the goal of providing an easy-to-use framework for benchmarking deep re-ID models and facilitating future research in re-ID. This library is written in Python, with some code based on Cython for optimisation and acceleration\footnote{Notably, with our Cython code the calculation of CMC ranks and mAP can be greatly accelerated, e.g., 3 mins shortened to 7 s on Market1501~\citep{zheng2015scalable}, 30 mins shortened to 3 mins on MSMT17~\citep{wei2018person}.}, and is built on top of PyTorch for automatic differentiation and fast tensor computation on GPUs.

The overall structure is shown in Listing \ref{lst:lib_structure}. There are totally 6 modules among which we will discuss 3 main modules in more details in the next section, namely \texttt{data}, \texttt{engine} and \texttt{models}. The library is fully documented and is easy to install. It also hosts a model zoo with various pre-trained re-ID CNN models publicly available for download.

\begin{lstlisting}[language=sh, float=tp, caption={Structure of Torchreid library.}, label={lst:lib_structure}]
torchreid/
  data/ # data loaders, data augmentation methods, data samplers
  engine/ # training and evaluation pipelines
  losses/ # loss functions
  metrics/ # distance metrics, evaluation metrics
  models # CNN architectures
  optim/ # optimiser and learning rate schedulers
  utils/ # useful tools (also suitable for other PyTorch projects)
\end{lstlisting}

\begin{lstlisting}[language=python, float=tp, caption={Example of using the high-level APIs in Torchreid for model training and test.}, label={lst:highlevel_apis}]
# Step 1: import the Torchreid library
import torchreid

# Step 2: construct data manager
datamanager = torchreid.data.ImageDataManager(
    root='reid-data',
    sources='market1501',
    targets='market1501',
    height=256,
    width=128,
    batch_size_train=32,
    batch_size_test=100,
    transforms=['random_flip', 'random_crop']
)

# Step 3: construct CNN model
model = torchreid.models.build_model(
    name='resnet50',
    num_classes=datamanager.num_train_pids,
    loss='softmax',
    pretrained=True
)
model = model.cuda()

# Step 4: initialise optimiser and learning rate scheduler
optimizer = torchreid.optim.build_optimizer(
    model,
    optim='adam',
    lr=0.0003
)

scheduler = torchreid.optim.build_lr_scheduler(
    optimizer,
    lr_scheduler='single_step',
    stepsize=20
)

# Step 5: construct engine
engine = torchreid.engine.ImageSoftmaxEngine(
    datamanager,
    model,
    optimizer=optimizer,
    scheduler=scheduler,
    label_smooth=True
)

# Step 6: run model training and test
engine.run(
    save_dir='log/resnet50',
    max_epoch=60,
    eval_freq=10,
    print_freq=10,
    test_only=False
)
\end{lstlisting}

\section{Main Modules} \label{sec:main_modules}
\subsection{Data} \label{sec:data_loaders}
Perhaps during the implementation of a computer vision project (or more generally a machine learning project), the most effort is not devoted to the implementation of a particular model or training algorithm, but to the construction of unified data loaders for preparing data. In Torchreid, we build each dataset on top of base classes (\texttt{ImageDataset} and \texttt{VideoDataset}), which provide basic functions for sampling, reading and pre-processing images. Therefore, for a new dataset users only need to define its training, query and gallery image sets, more specifically, the image paths, identity and camera view labels. To allow seamless combination of different re-ID datasets, the base classes are implemented with a customised \texttt{add} function such that different dataset \emph{instances} can be directly summed up to obtain the combined dataset.

The training and test data loaders are wrapped in a high-level class called \texttt{DataManager}, which is responsible for the construction of sampling strategy, data augmentation methods and data loaders. An instance of \texttt{DataManager} will serve as the input to the training pipeline. We have \texttt{ImageDataManager} for image datasets and \texttt{VideoDataManager} for video datasets, both being the child classes of \texttt{DataManager}.

Currently, Torchreid supports the following re-ID datasets that are commonly used in the literature,
\begin{itemize}
    \item \textbf{Image datasets}: Market1501~\citep{zheng2015scalable}, CUHK03~\citep{li2014deepreid}, DukeMTMC-reID~\citep{ristani2016performance,zheng2017unlabeled}, MSMT17~\citep{wei2018person}, VIPeR~\citep{gray2007evaluating}, GRID~\citep{loy2009multi}, CUHK01~\citep{CUHK01}, SenseReID~\citep{zhao2017spindle}, QMUL-iLIDS~\citep{iLIDS}, PRID~\citep{PRID}, and CUHK02~\citep{CUHK02}.
    \item \textbf{Video datasets}: MARS~\citep{zheng2016mars}, iLIDS-VID~\citep{wang2014person}, PRID2011~\citep{hirzer2011person}, and DukeMTMC-VideoReID~\citep{ristani2016performance,wu2018exploit}.
\end{itemize}

All datasets are implemented with their \emph{de facto} evaluation protocols so that the evaluation results can be fairly compared with published papers.

Listing~\ref{lst:highlevel_apis} (step 2) shows an example of how to construct a \texttt{ImageDataManager} to do model training and evaluation on Market1501. Note that the \texttt{sources} and \texttt{targets} attributes can take any number of datasets as input, as long as the datasets are available. For example, when \texttt{sources=['market1501', 'dukemtmcreid']}, the training data will be the combined training images from Market1501 and DukeMTMC-reID; when \texttt{targets=['market1501', 'dukemtmcreid']}, it will return the query and gallery images from Market1501 and DukeMTMC-reID separately.

\subsection{Engine} \label{sec:engines}
The \texttt{engine} module, which aims to provide a streamlined pipeline for training and evaluation of deep re-ID models, is carefully designed to be as modular as possible for easy extension. Concretely, a generic base \texttt{Engine} is implemented for both image- and video-based re-ID to provide universal training loops and other reusable features, such as data parsing, model checkpointing and performance measurement. Therefore, new engines can subclass this \texttt{Engine} to reduce tedious code re-writing.

Based on the \texttt{Engine}, two learning paradigms are currently implemented for CNN model training, one is classification with \texttt{softmax} loss (\texttt{ImageSoftmaxEngine} \& \texttt{VideoSoftmaxEngine}) and the other is metric learning with \texttt{triplet} loss (\texttt{ImageTripletEngine} \& \texttt{VideoTripletEngine}). These two learning paradigms have been widely adopted in recent top-performing re-ID models~\citep{hermans2017defense,li2018harmonious,chang2018multi,zhou2019osnet,chen2019mixed,zhou2019learning}. An example of how to use this module is illustrated in steps 5 \& 6 of Listing~\ref{lst:highlevel_apis}.

Besides the basic features necessary to construct a full train-evaluation pipeline, Torchreid also provides some advanced training tricks to improve the re-ID performance. For instance, to reduce overfitting the label smoothing regulariser~\citep{szegedy2016rethinking} is implemented for the \texttt{softmax} pipeline; for better transfer learning the pipeline allows the pre-trained CNN layers to be frozen during early training~\citep{geng2016deep} where the layers are specified by users.

\begin{figure}[t]
    \centering
    \includegraphics[width=.98\textwidth]{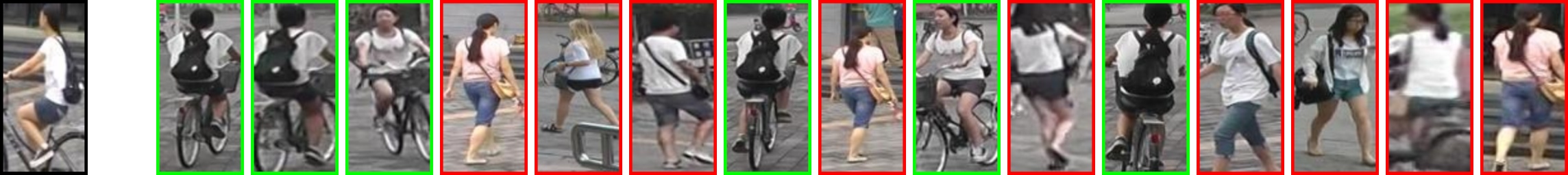}
    \caption{A visualised ranking list using Torchreid. The black, green and red colours outline the query image, correct matches and false matches, respectively.}
    \label{fig:ranked_results}
    \vspace{-0.3cm}
\end{figure}

\keypoint{Visualisation toolkit}
As qualitative result is often easier for us to understand how well a CNN model has learned, we implement two visualisation functions in Torchreid. The first function is \texttt{visrank}, which can visualise the ranking result of a re-ID CNN by saving for each query image the top-$k$ similar gallery images ($k$ is decided by users). Figure~\ref{fig:ranked_results} shows an example of the visualised ranking list.

\begin{wrapfigure}{r}{0.3\textwidth}
\centering
\includegraphics[width=0.29\textwidth]{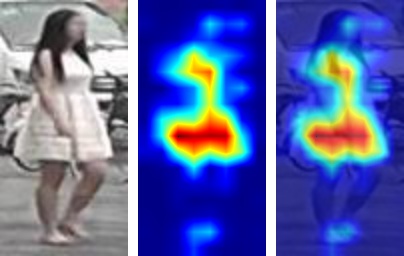}
\caption{Activation map.}
\label{fig:actmap}
\end{wrapfigure}

The second function is \texttt{visactmap}, which stands for visualising activation maps. Given an input image, the activation map can be used to analyse where the CNN focuses on to extract features~\citep{zhou2019osnet}. Specifically, an activation map is computed by taking the sum of absolute-valued feature maps (typically the highest-level feature maps) along the channel dimension, followed by a spatial $\ell$2 normalisation~\citep{zagoruyko2017paying}. An example is shown in Figure~\ref{fig:actmap}, which is obtained by OSNet~\citep{zhou2019osnet,zhou2019learning}. Intuitively, the image regions with warmer colours have higher activation values, which contribute the most to the generation of final feature vectors. Whereas the regions with cold colours are likely to contain less important/reliable regions for re-ID.

In addition, Torchreid is also integrated with PyTorch's built-in \texttt{SummaryWriter} for tensorboard\footnote{\url{https://www.tensorflow.org/tensorboard/}.} visualisation. For more features, please visit our github repository: \url{https://github.com/KaiyangZhou/deep-person-reid}.

\subsection{Models} \label{sec:models}
The \texttt{models} package provides a collection of implementations of state-of-the-art CNN architectures, which includes not only models that are specifically designed for re-ID, but also generic object recognition models that have been widely used as the re-ID CNN backbones.
The currently available models are listed below,
\begin{itemize}
    \item \textbf{ImageNet classification models}: ResNet~\citep{he2016deep}, ResNeXt~\citep{xie2017aggregated}, SENet~\citep{hu2018senet}, DenseNet~\citep{huang2017densely}, Inception-ResNet-V2~\citep{szegedy2017inception}, Inception-V4~\citep{szegedy2017inception}, and Xception~\citep{chollet2017xception}.
    \item \textbf{Lightweight models}: NASNet~\citep{NASNet}, MobileNetV2~\citep{sandler2018mobilenetv2}, ShuffleNet(V2)~\citep{zhang2018shufflenet,ma2018shufflenetv2}, and SqueezeNet~\citep{iandola2016squeezenet}.
    \item \textbf{Re-ID specific models}: MuDeep~\citep{qian2017multi}, ResNet-mid~\citep{yu2017devil}, HACNN~\citep{li2018harmonious}, PCB~\citep{sun2018beyond}, MLFN~\citep{chang2018multi}, OSNet~\citep{zhou2019osnet}, and OSNet-AIN~\citep{zhou2019learning}.
\end{itemize}

In addition to the CNN model code, we also release the corresponding model weights trained on the re-ID datasets (as well as the ImageNet pre-trained weights for some models) to further facilitate person re-ID research. We will keep updating this \texttt{models} package by incorporating new CNN models.

\section{Discussion} \label{sec:discussion}
Open-source frameworks are vital for pushing forward the progress of deep learning (DL) research. These include not only the backend engines for DL frameworks, such as PyTorch and TensorFlow, but also their extended libraries and toolkits that constitute the DL ecosystems, covering different research areas and providing convenient tools for fast research prototyping and development. In particular, recent years have witnessed an emergence of remarkable open-source frameworks/libraries developed for a wide range of research topics, e.g., Detectron~\citep{Detectron2018}, MMDetection~\citep{mmdetection} and SimpleDet~\citep{chen2019simpledet} for object detection, MMAction~\citep{mmaction2019} for action recognition, AllenNLP~\citep{gardner2018allennlp} and Transformers~\citep{Wolf2019HuggingFacesTS} for natural language processing, Torchmeta~\citep{deleu2019torchmeta} for meta-learning, etc.

These open-source projects can greatly reduce the efforts for researchers and practitioners to reproduce state-of-the-art models and moreover, provide streamlined pipelines for easy development and extension. With a similar goal, Torchreid is specifically designed for DL and person re-ID research. Though there exist some excellent open-source projects for re-ID, such as \texttt{Open-ReID}\footnote{\url{https://github.com/Cysu/open-reid}.} and \texttt{Person\_reID\_baseline\_pytorch}\footnote{\url{https://github.com/layumi/Person_reID_baseline_pytorch}.}, Torchreid is clearly different due to its unique features and versatility. In the future, we are committed to maintaining Torchreid and keeping it up-to-date by, for example, adding new datasets, CNN models and training algorithms.

{\small
\bibliographystyle{apalike}
\bibliography{reference}
}

\end{document}